\documentclass[10pt,twocolumn,letterpaper]{article}

\usepackage{cvpr}              
\definecolor{cvprblue}{rgb}{0.21,0.49,0.74}
\usepackage[pagebackref,breaklinks,colorlinks,allcolors=cvprblue]{hyperref}
\usepackage{comment}
\usepackage{multirow}
\usepackage{caption}

\def\paperID{18564} 
\def\confName{CVPR}
\def\confYear{2026}

\title{CUS-GS: A Compact Unified Structured Gaussian Splatting Framework for Multimodal Scene Representation}

\author{
Yuhang Ming$^1$, Chenxin Fang$^1$, Xingyuan Yu$^2$, Fan Zhang$^3$, Weichen Dai$^1$, \\
Wanzeng Kong$^{1*}$, Guofeng Zhang$^2$
\\
$^1$School of Computer Science, Hangzhou Dianzi University, $^2$CAD \& CG, Zhejiang University, \\
$^3$School of Computer Science, University of Bristol\\
{\tt\small \{yuhang.ming, chenxin.fang, daiweichen, kongwanzeng\}@hdu.edu.cn,} \\
{\tt\small \{rickyyxy, zhangguofeng\}@zju.edu.cn, fan.zhang@bristol.ac.uk}\\
{\small $^*$Corresponding author.}
}

\begin{document}

\twocolumn[{
\maketitle
\vspace{-5ex}
\begin{center}
    \includegraphics[width=\linewidth]{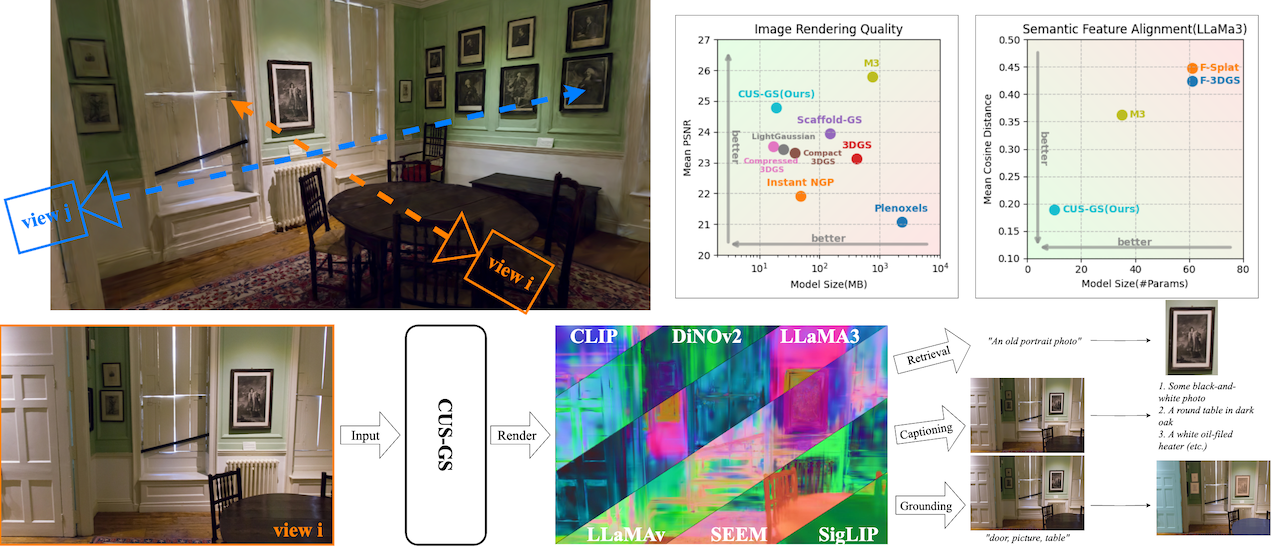}
\end{center}
\vspace{-0.5cm}
\captionsetup{type=figure}
\captionof{figure}{%
    \textbf{CUS-GS} is the \emph{first} framework to unify structured 3DGS with multimodal semantic modeling. The voxel-anchored structured design produces a geometry-aware and multimodally aligned 3D feature field, while maintaining high efficiency—achieving competitive performance with as few as \emph{6M} parameters, comparing to \emph{35M} of the closest rival.
}\label{fig:teaser}
\vspace{0.2cm}
}]

\begin{abstract}
Recent advances in Gaussian Splatting based 3D scene representation have shown two major trends: semantics-oriented approaches that focus on high-level understanding but lack explicit 3D geometry modeling, and structure-oriented approaches that capture spatial structures yet provide limited semantic abstraction. To bridge this gap, we present \textbf{CUS-GS}, a compact unified structured Gaussian Splatting representation, which connects multimodal semantic features with structured 3D geometry. Specifically, we design a voxelized anchor structure that constructs a spatial scaffold, while extracting multimodal semantic features from a set of foundation models (e.g., CLIP, DINOv2, SEEM). Moreover, we introduce a multimodal latent feature allocation mechanism to unify appearance, geometry, and semantics across heterogeneous feature spaces, ensuring a consistent representation across multiple foundation models. Finally, we propose a feature-aware significance evaluation strategy to dynamically guide anchor growing and pruning, effectively removing redundant or invalid anchors while maintaining semantic integrity. Extensive experiments show that CUS-GS achieves competitive performance compared to state-of-the-art methods using as few as 6M parameters — an order of magnitude smaller than the closest rival at 35M — highlighting the excellent trade off between performance and model efficiency of the proposed framework.
\end{abstract}    
\section{Introduction}
\label{sec:intro}

3D scene representation has long been a central topic in computer vision and robotics. As a fundamental module, it bridges perception~\cite{mascaro_scene_2025_annurev, ming_benchmarking_2025_eaai} with various downstream tasks such as 3D reconstruction~\cite{zhi_scenecode_2019_cvpr, li_voxsurf_2024_tvcg}, SLAM~\cite{matsuki_monogs_2024_cvpr, ming_slc2slam_2025_ral}, navigation~\cite{wang_navigation_2021_cvpr, wang_gridmm_2023_cvpr}, and manipulation~\cite{nair_r3m_2022_corl, yu_artgs_2025_iros}.
In recent years, the community has witnessed remarkable progress in neural rendering–based scene representations (e.g., NeRF~\cite{mildenhall_nerf_2021_commun} and 3DGS~\cite{kerbl_3dgs_2023_tog}), providing a powerful differentiable framework. Among them, 3DGS is particularly promising for its high-fidelity real-time rendering capabilities.

To accommodate different downstream tasks, recent studies on 3DGS–based scene representations have diversified into multiple directions. Among them, two prominent trends are particularly relevant to this work.
The first emphasizes structural hierarchy and rendering efficiency, forming the structure-oriented branch. Representative works, such as Scaffold-GS~\cite{lu_scaffoldgs_2024_cvpr} and Octree-GS~\cite{ren_octreegs_2025_pami}, enhance spatial compactness through voxelization and adaptive spatial subdivision.
The second line of research focuses on semantics-oriented extensions, integrating multimodal knowledge and high-level understanding into 3DGS. Examples include Feature 3DGS~\cite{zhou_feature3dgs_2024_cvpr} and M3~\cite{zou_m3spatial_2025_iclr}.
Despite these remarkable advancements, a notable gap still remains between these two paradigms: structure-oriented methods lack semantic understanding capabilities, whereas semantics-oriented methods often compromise spatial consistency and structural regularity.

Targeting this gap, we present CUS-GS (shown in ~\autoref{fig:teaser}), a Compact Unified Structured Gaussian Splatting framework for multimodal 3D scene representation. It aims to endow Gaussian splatting based scene representations with high-level multimodal understanding while maintaining a compact and efficient model structure. In particular, instead of treating structure and semantics as separate objectives, our CUS-GS integrates them within a unified representation that preserves spatial structure while remaining semantically expressive. 

To this end, we introduce \emph{a structured multimodal scene representation}, composed of a 3D voxelized anchor scaffold and a multimodal feature memory. Specifically, we first construct the multimodal memory by extracting features from a set of foundation models and applying redundancy reduction within each feature space. Each voxel anchor maintains a learnable latent feature that governs $N$ 3D Gaussians, which are decoded via a set of shared MLPs to recover their 3D Gaussian attributes and multimodal semantic features. To ensure robust decoding across heterogeneous feature spaces, we design a multimodal latent feature allocation mechanism together with a \emph{hierarchical query adaptation} module, enabling a consistent representation of appearance, geometry, and semantics. Finally, we propose a \emph{feature-aware anchor pruning} strategy that enforces compactness by jointly considering the collective significance of the Gaussians governed by each anchor, as well as the capacity and learning dynamics of its latent features. The primary contributions of our proposed CUS-GS are summarized as follows:

\begin{enumerate}

\item This is the \textbf{first attempt} to combine structure and semantics within a \textbf{unified}, \textbf{compact}, and \textbf{efficient} 3DGS framework for \textbf{multimodal scene understanding}.



\item The novel \textbf{hierarchical query adaptation mechanism} propagates each anchor’s latent feature to its associated Gaussian queries through shared MLPs, which significantly reduces the model size (compared to M3 \cite{zou_m3spatial_2025_iclr} while achieving spatially coherent and semantically adaptive rendering.

\item The new \textbf{feature-aware anchor pruning strategy} jointly considers the rendering-based significance of the Gaussians governed by each anchor and the capacity and learning dynamics of its latent features, rather than performing pruning based on Gaussian attributes~\cite{lu_scaffoldgs_2024_cvpr, fan_lightgaussian_2024_nips} - this enables a compact yet semantics-preserving structured representation.
\end{enumerate}

We conduct extensive experiments on several benchmarks to evaluate the performance of our proposed CUS-GS. Results show that CUS-GS achieves competitive image rendering quality and strong multimodal feature alignment across multiple foundation models, while maintaining a model size up to \textbf{5× smaller} than state-of-the-art methods. Furthermore, experiments on high-level downstream tasks demonstrate that CUS-GS effectively balances semantic consistency, structural fidelity, and model compactness, highlighting its advantage as a unified and efficient multimodal scene representation framework.


\section{Related Work}
\label{sec:related}

In this section, we review related works on neural rendering–based 3D scene representation.
Given the rapid progress and vast literature in this area, we focus on the studies most relevant to our problem setting.
We begin with a brief overview of neural rendering–based 3D scene representations, followed by two trending directions: structure-oriented 3DGS and semantic-oriented 3DGS.
For a more comprehensive review, readers are referred to existing surveys~\cite{tewari_neuralrendering_2020_cgf, tewari_neuralrendering_2022_cgf, liao_nerf_2025_compsurvey, ming_benchmarking_2025_eaai, bao_3dgs_2025_tcsvt, bagdasarian_3dgszip_2025_cgf}.

\vspace{4pt}\noindent\textbf{Neural Rendering-based Scene Representation:}
Neural rendering, as the integration of traditional computer graphics and modern deep learning, has established a new paradigm for 3D scene representation~\cite{tewari_neuralrendering_2020_cgf, tewari_neuralrendering_2022_cgf}. 
With the emergence of NeRF~\cite{mildenhall_nerf_2021_commun}, neural rendering has rapidly become a leading topic in scene representation and related fields, inspiring a wide range of neural implicit approaches~\cite{barron_mipnerf360_2022_cvpr, tancik_blocknerf_2022_cvpr, albert_dnerf_2021_cvpr, muller_instancengp_2022_tog, li_vqdvgo_2023_cvpr, fridovich_plenoxel_2022_cvpr}. 
These methods offer high-fidelity novel view synthesis, but struggle with real-time rendering.
To address this limitation, many studies have turned to explicit neural primitives, with 3DGS~\cite{kerbl_3dgs_2023_tog} becoming a central framework. 
Subsequent work has advanced 3DGS along multiple directions — including expressive capability~\cite{chen_pgsr_2025_tvcg, xie_unigs_2025_arxiv}, robustness~\cite{wu_4dgs_2024_cvpr, kerbl_hierarchicalgs_2024_tog}, efficiency~\cite{fan_lightgaussian_2024_nips, lee_compactgs_2024_cvpr, niedermayr_compressed3dgs_2024_cvpr}, and generalizability~\cite{zhang_transplat_2025_aaai, zhou_gpsgaussianp_2025_pami}. 

\vspace{4pt}\noindent\textbf{Structure-oriented 3DGS:}
Targeting spatial compactness, a series works have investigated the incorporation of spatial structures and hierarchical organizations. 
As one of the earliest attempts, Scaffold-GS~\cite{lu_scaffoldgs_2024_cvpr} divides the 3D space into a scaffold of voxels and uses voxel anchors to manage nearby 3D Gaussians. Following this idea, various attempts have been explored to make the spatial distribution of 3D Gaussian more controllable and compact. Examples include 2D grids in self-organizing gaussian grids~\cite{morgenstern_2dgrids_2024_eccv}, 3D cubes in GaussianCube~\cite{zhang_gaussiancube_2024_nips}, octrees in OG-Mapping~\cite{wang_ogmapping_2024_arxiv}, Octree-GS~\cite{ren_octreegs_2025_pami}, GS-Octree~\cite{li_gsoctree_2024_arxiv}, and hash-grids in HAC~\cite{chen_hac_2024_eccv} and HAC++~\cite{chen_hacpp_2025_pami}. 
Alternatively, other works such as Hierarchical 3DGS~\cite{kerbl_hierarchicalgs_2024_tog}, HUG~\cite{su_hug_2025_iccv}, and Virtualized 3D Gaussians~\cite{yang_virtualized_2025_siggraph} organize Gaussians in hierarchical blocks or clusters, enabling level-of-detail rendering for large-scale environments. 
Meanwhile, RGBDS-SLAM~\cite{cao_rgbdsslam_2025_ral}, HiSplat~\cite{tang_hisplat_2025_iclr}, HRGS~\cite{li_hrgs_2025_arxiv}, and PyGS~\cite{wang_pygs_2024_arxiv} construct pyramidal or coarse-to-fine hierarchies, supporting multi-scale reconstruction and high-resolution rendering.

\vspace{4pt}\noindent\textbf{Semantic-oriented 3DGS:}
Aiming to make 3DGS multimodally expressive, a parallel line of works augments 3DGS with semantic features extracted from foundation models.
Early approaches distill 2D semantic features from models like CLIP~\cite{radford_clip_2021_icml}, DiNOv2~\cite{oquab_dinov2_2024_arxiv}, and SAM~\cite{kirillov_sam_2023_iccv} into the 3DGS representation, as seen by Feature 3DGS~\cite{zhou_feature3dgs_2024_cvpr}, FMGS~\cite{zuo_fmgs_2025_ijcv}, Semantic Gaussians~\cite{guo_semanticgaussian_2024_arxiv}, LEGaussians~\cite{shi_legaussians_2024_cvpr}, and Feature-Splat~\cite{qiu_featuresplat_2024_eccv}. 
Concurrently, other works including LUDVIG~\cite{marrie_ludvig_2025_iccv}, SLAG~\cite{szilagyi_slag_2025_ral}, and Dr. Splat~\cite{kim_drsplat_2025_cvpr} directly project foundation model features onto the Gaussians, avoiding additional training modules and improving scalability.. 
In parallel, CLIP-GS~\cite{jiao_clipgs_2025_iccv}, UniGS~\cite{li_unigs_2025_iclr}, and M3~\cite{zou_m3spatial_2025_iclr} introduce trainable latent features to each 3D Gaussian and align images, text, and 3D into a shared embedding space. 

\section{Methodology}
\label{sec:method}

\begin{figure*}[t]
    \centering
    \includegraphics[width=\linewidth]{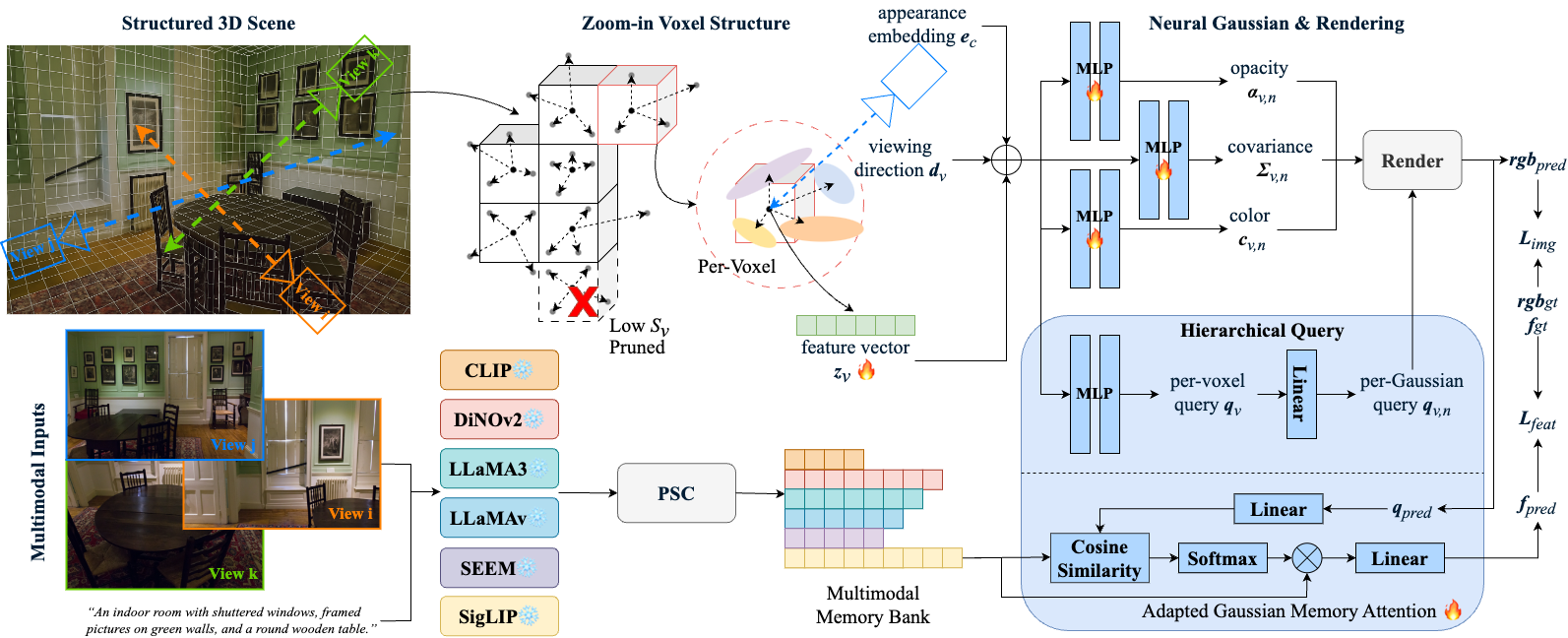}
    \caption{\textbf{Architecture Overview.} 
     Our CUS-GS bridges structured 3DGS with multimodal scene understanding through a voxelized scaffold and a unified multimodal memory bank.
    Each voxel maintains a latent feature that, together with view-dependent appearance cues, is decoded into hierarchical queries and Gaussian attributess.
    Multimodal features extracted from 6 foundation models are first compressed via PSC, and are then attended by learned queries to retrieve aligned semantic features.
    The resulting Gaussian attributes drive differentiable splatting, while the queried semantics enrich the representation, yielding compact, spatially consistent, and multimodally expressive 3D scene representations.
    }
    \label{fig:architecture}
\end{figure*}

3DGS is developing rapidly, with numerous works advancing along two major directions — structure-oriented and semantic-oriented representations. 
Our approach integrates these two perspectives by constructing a unified structured multimodal representation that jointly encodes spatial layout and multimodal features from six foundation models.
The overall architecture of our proposed CUS-GS framework is illustrated in \autoref{fig:architecture}.
In the following, we first briefly review the background of 3DGS and the principal scene components in \cref{sec::prelim} to make the paper self-contained. We then present our method in detail — \cref{sec::structured_rep} introduces the structured multimodal representation, \cref{sec::hierarhical_query} describes the hierarchical query adaptation,  \cref{sec::grow_prune} explains the feature-aware growing and pruning strategy, and \cref{sec::loss} provides training details.

\subsection{Preliminaries}
\label{sec::prelim}

\noindent\textbf{3D Gaussian Splatting (3DGS)}~\cite{kerbl_3dgs_2023_tog} models a 3D scene using a collection of \emph{spatially independent} anisotropic Gaussian primitives ${G_i}_{i=1}^N$.
Each Gaussian $G_i$ is parameterized by a mean $\boldsymbol{\mu}_i \in \mathbb{R}^3$, a covariance matrix $\boldsymbol{\Sigma}_i \in \mathbb{R}^{3\times3}$, 
an opacity $\alpha_i \in [0,1]$, and view-dependent color coefficients $\boldsymbol{c}_i$ modeled by spherical harmonics. 

Given a camera projection matrix $\boldsymbol{P}$, each Gaussian is projected onto the image plane as a 2D Gaussian with mean 
$\boldsymbol{\mu}'_i = \boldsymbol{P}\boldsymbol{\mu}_i$ and covariance
\begin{equation}
\boldsymbol{\Sigma}'_i = \boldsymbol{J}_i \boldsymbol{\Sigma}_i \boldsymbol{J}_i^\top,
\end{equation}
where $\boldsymbol{J}_i$ denotes the Jacobian of the projection at $\boldsymbol{\mu}_i$. 
The rendered color at pixel $\boldsymbol{u}$ is then obtained by front-to-back $\alpha$-compositing of all projected Gaussians:
\begin{equation}
\boldsymbol{C}(\boldsymbol{u}) = \sum_{i=1}^{N} T_i \, \alpha_i \, \boldsymbol{c}_i \, 
\exp\!\left(-\tfrac{1}{2}(\boldsymbol{u}-\boldsymbol{\mu}'_i)^\top {\boldsymbol{\Sigma}'_i}^{-1} (\boldsymbol{u}-\boldsymbol{\mu}'_i)\right),
\end{equation}
where $T_i = \prod_{j<i} (1-\alpha_j)$ denotes the accumulated transmittance of all previous Gaussians along the ray. 

This differentiable rendering formulation allows 3DGS to be optimized end-to-end via standard photometric losses, 
providing an explicit and efficient representation for high-quality novel view synthesis.

\vspace{4pt}\noindent\textbf{Principal Scene Component (PSC)}~\cite{zou_m3spatial_2025_iclr} is a shared multimodal memory bank from multiple foundation models before linking it to 3D Gaussians. 
Given a set of scene views $\{I_v\}$ and a collection of foundation models $\{\mathcal{F}^j\}_{j=1}^J$, 
it first extracts dense feature maps from each model and granularity, and flattens them into a set of raw feature vectors
\[
\mathcal{R} = \{\boldsymbol{r}_n\}_{n=1}^{N_\mathrm{raw}}, \quad \|\boldsymbol{r}_n\|_2 = 1.
\]

To reduce redundancy while preserving the original embedding spaces, PSC applies a similarity-based reduction. In particular, it scans $\mathcal{R}$ sequentially and selects a vector $\boldsymbol{r}_n$ as a memory entry if its cosine similarity to all previously selected entries is below a threshold $\gamma$.
Formally, the resulting multimodal memory bank is
\[
\mathcal{M} = \{\boldsymbol{m}_k\}_{k=1}^K \subset \mathcal{R},
\quad 
\text{s.t.}\;
\max_{j<k} \; \boldsymbol{m}_k^\top \boldsymbol{m}_j < \gamma,
\]
where each $\boldsymbol{m}_k$ is directly taken from the original foundation-model features and optionally tagged with its source model and granularity. 
This memory bank provides a compact yet faithful collection of multimodal scene descriptors, which can later be queried by 3D representations.

\subsection{Structured Multimodal Representation}
\label{sec::structured_rep}

To organize both geometry and multimodal semantics in a unified manner, 
we divide the 3D space into a voxel scaffold with cell size $l$. 
Each voxel $v$ serves as a spatial anchor that governs the attributes of $N$ Gaussian primitives 
$\{G_{v,n}\}_{n=1}^{N}$ located within its spatial extent. 
Following the design philosophy of structural 3DGS methods~\cite{lu_scaffoldgs_2024_cvpr, ren_octreegs_2025_pami, chen_hac_2024_eccv, chen_hacpp_2025_pami}, we associate each voxel with a learnable feature vector 
$\boldsymbol{z}_v \in \mathbb{R}^{d_f}$ that jointly encodes its local appearance, geometric, and semantic context, while also interacting with view-related cues to achieve consistent rendering. 
Specifically, a view-dependent appearance embedding $\boldsymbol{e}_c \in \mathbb{R}^{d_c}$ is introduced for each camera $c$ to compensate for illumination and tone variations across views, and the voxel’s viewing direction $\boldsymbol{d}_v$ is incorporated to maintain angular coherence.

For each Gaussian inside voxel $v$, its attributes are decoded from the combination of 
$(\boldsymbol{z}_v, \boldsymbol{d}_v, \boldsymbol{e}_c)$ through a set of shared lightweight MLPs:
\begin{align}
\alpha_{v,n} &= \mathrm{MLP}_{\text{opacity}}(\boldsymbol{z}_v, \boldsymbol{d}_v, \boldsymbol{e}_c), \\
\boldsymbol{\Sigma}_{v,n} &= \mathrm{MLP}_{\text{cov}}(\boldsymbol{z}_v, \boldsymbol{d}_v, \boldsymbol{e}_c), \\
\boldsymbol{c}_{v,n} &= \mathrm{MLP}_{\text{color}}(\boldsymbol{z}_v, \boldsymbol{d}_v, \boldsymbol{e}_c),
\end{align}
where each network is shared across all voxels and jointly optimized during training. 
To maintain spatial flexibility, each Gaussian’s local position is parameterized by a learnable offset 
$\Delta\boldsymbol{x}_{v,n}$ relative to the voxel center.

In addition to geometric and appearance attributes, 
we further decode per-Gaussian semantic queries for multimodal reasoning. 
A separate network $\mathrm{F}_{\theta}$ maps the same input triplet: 
$(\boldsymbol{z}_v, \boldsymbol{d}_v, \boldsymbol{e}_c)$ to semantic query vectors:
\begin{equation}
\boldsymbol{q}_{v,n} = \mathrm{F}_{\theta}(\boldsymbol{z}_v, \boldsymbol{d}_v, \boldsymbol{e}_c),
\end{equation}
which serves as the intermediate representation for hierarchical query adaptation 
and multimodal memory retrieval introduced in Sec.~\ref{sec::hierarhical_query}.

This design allows each voxel to encapsulate local structure and multimodal context,
producing a compact and spatially coherent latent scaffold that can be consistently decoded 
into geometry, appearance, and semantic attributes across views.

\subsection{Hierarchical Query Adaptation}
\label{sec::hierarhical_query}

We retrieve multimodal semantic features through a query–memory mechanism similar to M3~\cite{zou_m3spatial_2025_iclr}, using PSCs as the memory bank (see \cref{sec::prelim}).
However, flat per-Gaussian querying often leads to redundant and spatially inconsistent retrievals,
as neighboring Gaussians may correspond to a similar visual context~\cite{zou_m3spatial_2025_iclr}.
To overcome this limitation, we introduce a hierarchical adaptation mechanism that propagates multimodal queries
from voxel-level anchors to individual Gaussians, ensuring both spatial coherence and semantic adaptability.

Specifically, each voxel feature $\boldsymbol{z}_v$ is first mapped to a voxel-level query $q_v \in \mathbb{R}^{d_q}$ through a lightweight MLP:
\begin{equation}
\boldsymbol{q}_v = \mathrm{MLP}_{qv}(\boldsymbol{z}_v, \boldsymbol{d}_v, \boldsymbol{e}_c),
\end{equation}
which encodes aggregated appearance, geometric, and multimodal cues within the voxel. 
For each Gaussian $G_{v,n}$ governed by voxel $v$, a finer query $q_{v,n} \in \mathbb{R}^{d_q}$ is derived from the voxel query and the Gaussian’s relative offset:
\begin{equation}
\boldsymbol{q}_{v,n} = \mathrm{Linear}_{qg}\!\left([\boldsymbol{q}_v, \Delta\boldsymbol{x}_{v,n}]\right),
\end{equation}
where $\Delta\boldsymbol{x}_{v,n}$ denotes the offset. 
This hierarchical propagation allows local Gaussian queries to inherit the semantic priors of their voxel anchors while capturing position-specific variation.

The per-Gaussian queries are first passed through standard rendering pipeline to obtain predicted queries $\boldsymbol{q}_{pred}$, which then attend to the multimodal memory bank via dot-product attention in the adapted gaussian memory attention module: 
\begin{equation}
a_{pred,k} = \frac{\exp(\boldsymbol{q}_{pred}^\top \boldsymbol{m}_k / \sqrt{d})}
{\sum_{j} \exp(\boldsymbol{q}_{v,n}^\top \boldsymbol{m}_j / \sqrt{d})},
\end{equation}
and the retrieved multimodal feature is obtained by weighted aggregation:
\begin{equation}
\tilde{\boldsymbol{f}}_{pred} = \sum_{k} a_{pred,k} \boldsymbol{m}_k.
\end{equation}
To handle distribution discrepancies among heterogeneous foundation-model features, 
a linear adaptation layer $\boldsymbol{W}_{\mathrm{adapt}}$ is further applied:
\begin{equation}
\boldsymbol{f}_{pred} = \boldsymbol{W}_{\mathrm{adapt}}\, \tilde{\boldsymbol{f}}_{pred}.
\end{equation}

Through this hierarchical query adaptation, we avoid storing high-dimensional queries for every Gaussian, achieving a more compact representation, while the shared voxel-level queries also encourage smoother semantics across neighboring Gaussians.

\subsection{Feature-aware Pruning}
\label{sec::grow_prune}

To enhance compactness and better exploit voxel-level information,
we introduce a feature-aware pruning strategy that evaluates the importance of each voxel using both voxel-feature statistics and Gaussian-attribute indicators.

For voxel-feature statistics, we use two measures:
the feature norm $|\boldsymbol{z}_v|2$, which reflects the magnitude of the learned representation,
and the gradient norm $\|\nabla_{\boldsymbol{z}_v}\mathcal{L}\|_2$, which indicates the remaining learning dynamics and thus the potential for further optimization.

For Gaussian-attribute indicators, 
we compute per-Gaussian contribution score $S_{v,n}^{\text{contrib}}$ following the global significance formulation in~\cite{fan_lightgaussian_2024_nips},
which accumulates its hit counts over training rays weighted by opacity and normalized volume.
We then aggregate the voxel-level scores as $S_v^{\text{contrib}} = \sum_{n=1}^{N} S_{v,n}^{\text{contrib}}$.

The final significance score for voxel $v$ is defined as
\begin{equation}
S_v = \lambda_{v,n} \|\boldsymbol{z}_v\|_2 + \lambda_{v,g} \|\nabla_{\boldsymbol{z}_v}\mathcal{L}\|_2 + S_v^{contrib}.
\end{equation}

As voxels with low $S_v$ are pruned, this feature-aware criterion adaptively refines the voxel scaffold by balancing representation strength, learning potential, and rendering contribution.

\subsection{Loss Design}
\label{sec::loss}
The overall training objective combines image reconstruction and multimodal feature alignment losses:
\begin{equation}
\mathcal{L} = \mathcal{L}_{img} + \lambda_{l,e}\,\mathcal{L}_{feat}.
\end{equation}
The image reconstruction loss $\mathcal{L}_{img}$ enforces photometric consistency between rendered and ground-truth RGB images, defined as
\begin{equation}
\mathcal{L}_{img} = \lambda_{l,n}\,\mathcal{L}_{L1} + \lambda_{l,ssim}\,\mathcal{L}_{SSIM} + \lambda_{l,s}\,\mathcal{L}_{scale},
\end{equation}
where $\mathcal{L}_{L1}$ and $\mathcal{L}_{SSIM}$ are pixel-wise $\ell_1$ and structural similarity losses, respectively, and $\mathcal{L}_{scale}$ is an anchor-scale regularization term to stabilize voxel growth.  

The multimodal embedding loss $\mathcal{L}_{feat}$ aligns the reconstructed features with their foundation-model counterparts:
\begin{equation}
\mathcal{L}_{feat} = \mathcal{L}_{\ell_2} + \mathcal{L}_{cos},
\end{equation}
where $\mathcal{L}_{\ell_2}$ mmatches feature magnitudes and $\mathcal{L}_{cos}$ enforces directional consistency.  
This combination ensures faithful RGB reconstruction while maintaining semantic consistency across multimodal embeddings.
\section{Experiments}
\label{sec:exp}

\subsection{Experimental Setup}

\textbf{Implementation Details.}
For the proposed CUS-GS framework, the voxel size is fixed to $l=0.01$ in all experiments. 
Each voxel governs $N=10$ Gaussians, and both the per-voxel feature vector and the per-camera appearance embedding are set to dimension $d_f = d_c = 32$. 
All MLPs consist of two layers with hidden dimensions equal to $d_f$ and ReLU activations. 
For feature-aware pruning, we remove anchors with the lowest $0.1\%$ importance and set $\lambda_{v,n}=2$ and $\lambda_{v,g}=8$, as validated in the ablation study.  

Regarding multimodal semantics, we follow M3~\cite{zou_m3spatial_2025_iclr} and use a $160-d$ query vector to encode features extracted from six foundation models: CLIP~\cite{radford_clip_2021_icml}, 
SigLIP~\cite{zhai_siglip_2023_iccv}, DINOv2~\cite{oquab_dinov2_2024_arxiv}, 
SEEM~\cite{zou_seem_2023_nips}, and LLaMA-3.1/3.2v~\cite{grattafiori_llama3_2024_arxiv}.  
We train all models for $30{,}000$ iterations using the Adam optimizer~\cite{kingma_adam_2015_iclr}, with the same learning rate schedule as M3~\cite{zou_m3spatial_2025_iclr},assigning different rates to appearance, geometry, and multimodal parameter groups.

\begin{table*}[t]
    \centering
    \caption{\textbf{Quantitative Comparison on Image Rendering Quality:}
    Our CUS-GS achieves competitive rendering quality with the \emph{smallest multimodal model size}, demonstrating the efficiency of our structured multimodal representation.}
     \resizebox{0.8\linewidth}{!}{
    \begin{tabular}{l|rccc|rccc} 
        \toprule
        \multirow{2}{*}{\textbf{Method}} & \multicolumn{4}{c|}{Mip-NeRF 360~\cite{barron_mipnerf360_2022_cvpr}} & \multicolumn{4}{c}{Tank and Temple~\cite{knapitsch_tankandtemples_2017_tog}} \\ 
        \cline{2-9} & Size$\downarrow$ & PSNR$\uparrow$ & SSIM$\uparrow$ & LPIPS$\downarrow$ & Size$\downarrow$ & PSNR$\uparrow$ & SSIM$\uparrow$ & LPIPS$\downarrow$ \\
        \hline
        Plenoxels~\cite{fridovich_plenoxel_2022_cvpr}     & 2.1GB & 23.08 & 0.626 & 0.463 & 2.3GB & 21.08 & 0.719 & 0.379 \\
        Instant NGP~\cite{muller_instancengp_2022_tog}      &  48MB & 25.59 & 0.699 & 0.331 &  48MB & 21.92 & 0.745 & 0.305 \\
        VQ-DVGO~\cite{li_vqdvgo_2023_cvpr}       &  63MB & 24.23 & 0.636 & 0.393 &     - &     - &     - &     - \\
        \hline
        3DGS~\cite{kerbl_3dgs_2023_tog}         & 734MB & \underline{27.21} & 0.815 & \underline{0.214} & 411MB & 23.14 & {0.841} & {0.183} \\
        Scaffold-GS~\cite{lu_scaffoldgs_2024_cvpr}   & 163MB & \textbf{28.84} & \textbf{0.848} & 0.220 & 148MB & {23.96} & \underline{0.853} & \underline{0.177} \\
        Compact 3DGS~\cite{lee_compactgs_2024_cvpr} &  48MB & 27.08 & 0.798 & 0.247 &  39MB & 23.32 & 0.831 & 0.201 \\
        Compressed 3DGS~\cite{niedermayr_compressed3dgs_2024_cvpr} & 28MB  & 27.03 & 0.802 & 0.238 & \textbf{17MB} & 23.54 & 0.838  & 0.189 \\
        LightGaussian~\cite{fan_lightgaussian_2024_nips} & \underline{45MB} & {27.13} & 0.806 & 0.237 &  25MB & 23.44 & 0.832 & 0.202 \\
        \hline
        M3$^{*}$~\cite{zou_m3spatial_2025_iclr}            & 1.1GB & 24.14 & 0.799 & 0.223 & 747MB & \textbf{25.80} & \textbf{0.875} & \textbf{0.171} \\
        \textbf{CUS-GS (Ours)}          &  \textbf{20MB} & 26.19 & \underline{0.822} & \textbf{0.193} &  \underline{19MB} & \underline{24.79} & 0.835 & 0.227 \\
        \bottomrule
    \end{tabular}
    }
    \label{table:img_render}
\end{table*}

\vspace{4pt}\noindent\textbf{Datasets and Metrics.} 
To evaluate CUS-GS, we conduct experiments on 13 scenes from three public datasets~\cite{barron_mipnerf360_2022_cvpr, knapitsch_tankandtemples_2017_tog, hedman_deepblending_2018_tog}. 
Following standard practice in 3DGS and multimodal scene representation~\cite{lu_scaffoldgs_2024_cvpr, fan_lightgaussian_2024_nips, zou_m3spatial_2025_iclr}, 
we adopt the same evaluation protocol as prior studies.
Ourbenchmark includes all nine real-world scenes from Mip-NeRF360~\cite{barron_mipnerf360_2022_cvpr}, (five outdoor and four indoor), the ``Train'' and ``Truck'' scenes from the Tank \& Temples~\cite{knapitsch_tankandtemples_2017_tog} dataset, and the ``Playroom'' and ``DrJohnson'' scenes from DeepBlending~\cite{hedman_deepblending_2018_tog}.

For evaluating imgage rendering quality, we adopt the standard metrics used in novel view synthesis: 
SSIM, PSNR, and LPIPS. 
To assess the quality of multimodal semantic features, we evaluate from both \emph{low-level feature alignment} and \emph{high-level downstream task performance} perspectives. 
For low-level feature evaluation, we compute cosine distance and $\ell_2$ distance between reconstructed and reference feature maps. 
For high-level evaluation, following M3~\cite{zou_m3spatial_2025_iclr} and Feature 3DGS~\cite{zhou_feature3dgs_2024_cvpr}, 
we report mIoU, cIoU, AP50, and AP60 for semantic retrieval with CLIP~\cite{radford_clip_2021_icml} features, 
and I2T@1/@5/@10 and T2I@1/@5/@10 for image–text retrieval with SigLIP~\cite{zhai_siglip_2023_iccv} features. 
We also report model size for all the experiments, including both checkpoint storage size and total parameter count.
All comparison results follow the numbers reported in the original papers; when unavailable, we reproduce them with the authors’ official implementations and mark them with an asterisk ($*$).

\vspace{4pt}\noindent\textbf{Baselines.}
Since we evaluate our proposed CUS-GS for both image rendering quality and semantic feature alignment, 
different sets of comparison methods are selected for each task. 
For color rendering, we include three NeRF-style methods,
Plenoxels~\cite{fridovich_plenoxel_2022_cvpr}, Instant NGP~\cite{muller_instancengp_2022_tog}, 
and VQ-DVGO~\cite{li_vqdvgo_2023_cvpr} - 
as well as five 3DGS-based approaches, Scaffold-GS~\cite{lu_scaffoldgs_2024_cvpr},
Compressed 3DGS~\cite{niedermayr_compressed3dgs_2024_cvpr}, Compact 3DGS~\cite{lee_compactgs_2024_cvpr}, and LightGaussian~\cite{fan_lightgaussian_2024_nips}, 
together with the original 3DGS~\cite{kerbl_3dgs_2023_tog}. 
For semantic evaluation, we further compare against foundation-model–enhanced 3DGS methods: 
Feature Splatting (F-Splat)~\cite{qiu_featuresplat_2024_eccv}, 
Feature 3DGS (F-3DGS)~\cite{zhou_feature3dgs_2024_cvpr}, 
and M3~\cite{zou_m3spatial_2025_iclr}, with M3 evaluated for both color rendering and multimodal semantics.

\begin{table*}[t]
    \caption{\textbf{Quantitative Comparison on Semantic Feature Alignment:} Our CUS-GS achieves competitive alignment with 5$\times$ fewer parameters. ``D.'' denotes Dataset scenes (T.: Train, G.: Garden, D.: DrJohnson, P.: Playroom), and ``\# P.'' the number of parameters.}
    \resizebox{1.\linewidth}{!}{
        \begin{tabular}{c|l|r|cc|cccc|cccc|cc}
            \hline
            \multirow{2}{*}{\textbf{D.}} & \multirow{2}{*}{\textbf{Method}} & \multirow{2}{*}{\textbf{\# P.}} & \multicolumn{2}{c|}{\textbf{DiNOv2}} & \multicolumn{2}{c}{\textbf{CLIP}} & \multicolumn{2}{c|}{\textbf{SigLIP}} & \multicolumn{2}{c}{\textbf{SEEM}} & \multicolumn{2}{c|}{\textbf{LLaMA3}} & \multicolumn{2}{c}{\textbf{LLaMAv}} \\
            & & & Cosine$\downarrow$ & $\ell_2$$\downarrow$& Cosine$\downarrow$  & $\ell_2$$\downarrow$ & Cosine$\downarrow$  & $\ell_2$$\downarrow$ & Cosine$\downarrow$ & $\ell_2$$\downarrow$& Cosine$\downarrow$ & $\ell_2$$\downarrow$ & Cosine$\downarrow$ & $\ell_2$$\downarrow$ \\
            \hline
            \multirow{4}{*}{T.}
            & F-Splat~\cite{qiu_featuresplat_2024_eccv} &   61M & 0.6833 & 1.9835 & 0.5998 & 0.4779 & 0.6346 & 0.7851 & 0.4269 & 11.720 & 0.5300 & 0.2900 & 0.7026 & 56.23 \\
            & F-3DGS~\cite{zhou_feature3dgs_2024_cvpr}  &   61M & 0.3790 & 1.0108 & 0.3330 & 0.1540 & 0.3692 & 0.3328 & 0.1063 & 0.1034 & 0.4993 & 0.0150 & 0.6288 & 46.48 \\
            & M3~\cite{zou_m3spatial_2025_iclr}      &   35M & 0.5321 & 1.6810 & 0.3140 & 0.2800 & 0.2811 & 0.5096 & 0.1389 & 0.2251 & 0.4401 & 0.0253 & 0.7069 & 53.43 \\ 
            & CUS-GS (Ours)    &   13M & 0.5515 & 1.7454 & 0.3254 & 0.2871 & 0.2905 & 0.5237 & 0.2036 & 0.2942 & 0.2331 & 0.0251 & 0.7196 & 53.56 \\
            \hline
            \multirow{4}{*}{G.}
            & F-Splat\cite{qiu_featuresplat_2024_eccv} &   61M & 0.7328 & 1.9567 & 0.7005 & 1.3570 & 0.7247 & 0.8698 & 0.4224 & 9.4675 & 0.4944 & 0.3314 & 0.7443 & 60.83 \\ 
            & F-3DGS~\cite{zhou_feature3dgs_2024_cvpr}  &   61M & 0.2295 & 0.6033 & 0.2105 & 0.0945 & 0.2697 & 0.2585 & 0.1071 & 0.1424 & 0.4139 & 0.0141 & 0.4913 & 43.08 \\ 
            & M3~\cite{zou_m3spatial_2025_iclr}      &   35M & 0.5701 & 1.7279 & 0.3168 & 0.2876 & 0.2927 & 0.0004 & 0.1839 & 0.3469 & 0.3387 & 0.0217 & 0.7235 & 58.04 \\
            & CUS-GS (Ours)    &   13M & 0.6060 & 1.7991 & 0.3349 & 0.2980 & 0.3056 & 0.5710 & 0.1965 & 0.2817 & 0.1699 & 0.0181 & 0.7374 & 58.03 \\
            \hline
            \multirow{4}{*}{D.}                              
            & F-Splat~\cite{qiu_featuresplat_2024_eccv} &   61M & 0.8107 & 2.0333 & 0.6689 & 0.7877 & 0.6826 & 0.7744 & 0.4650 & 10.411 & 0.3757 & 0.0145 & 0.8184 & 54.82 \\
            & F-3DGS~\cite{zhou_feature3dgs_2024_cvpr}  &   61M & 0.4190 & 1.1279 & 0.3344 & 0.1537 & 0.3846 & 0.3552 & 0.1693 & 0.2169 & 0.3853 & 0.0150 & 0.6669 & 47.35 \\
            & M3~\cite{zou_m3spatial_2025_iclr}      &   35M & 0.5878 & 1.7553 & 0.3435 & 0.2924 & 0.2975 & 0.5366 & 0.2456 & 0.4179 & 0.3175 & 0.0226 & 0.7224 & 52.68 \\
            & CUS-GS (Ours)    &    8M & 0.6063 & 1.7851 & 0.3528 & 0.2987 & 0.3060 & 0.5503 & 0.2063 & 0.2864 & 0.1707 & 0.0182 & 0.7602 & 54.18 \\
            \hline
            \multirow{4}{*}{P.}
            & F-Splat~\cite{qiu_featuresplat_2024_eccv} &   61M & 0.7956 & 1.9640 & 0.6458 & 0.7808 & 0.6839 & 0.7678 & 0.4745 & 10.873 & 0.3915 & 0.0136 & 0.8185 & 59.42 \\ 
            & F-3DGS~\cite{zhou_feature3dgs_2024_cvpr}  &   61M & 0.4867 & 1.2193 & 0.3813 & 0.1726 & 0.4571 & 0.4094 & 0.1714 & 0.2103 & 0.3987 & 0.0139 & 0.6922 & 52.50 \\ 
            & M3~\cite{zou_m3spatial_2025_iclr}      &   35M & 0.6074 & 1.7545 & 0.3260 & 0.2987 & 0.2951 & 0.5623 & 0.2560 & 0.4584 & 0.3555 & 0.0241 & 0.7288 & 57.38 \\
            & CUS-GS (Ours)    &    6M & 0.6156 & 1.7782 & 0.3345 & 0.3050 & 0.3041 & 0.5718 & 0.2126 & 0.2937 & 0.1810 & 0.0188 & 0.7333 & 57.10 \\
            \hline
        \end{tabular}
    }
    \label{table:feat_align}
\end{table*}

\subsection{Experimental Results}

\noindent\textbf{Quantitative Results.}
\autoref{table:img_render} presents the quantitative results on image rendering quality. 
Across all datasets, CUS-GS maintains the smallest model size (20–19 MB), more than an order of magnitude smaller than M3 (1.1 GB) and comparable to the most compact appearance-only 3DGS variants. 
On the Mip-NeRF360 dataset, our method achieves the best LPIPS (0.193) and the second-highest SSIM (0.822), 
reflecting strong perceptual and structural fidelity despite a moderate PSNR drop. This trend is consistent with the dataset’s diverse, texture-rich scenes, where multimodal supervision provides meaningful regularization and reduces sensitivity to pixel-level noise.
On the geometry-dominant Tank and Temples dataset, CUS-GS attains the second-best PSNR (24.79) while retaining competitive SSIM and LPIPS scores, 
indicating that the model preserves high-quality geometric reconstruction even with multimodal feature integration.

\autoref{table:feat_align} summarizes the quantitative comparison on semantic feature alignment. 
Overall, CUS-GS performs on par with M3 across most foundation models, 
while showing a clear and consistent improvement on LLaMA3 features. 
This suggests that the voxel-structured representation and hierarchical query adaptation offer stronger spatial support for language-driven semantic embeddings.
A slight decrease is observed for DINOv2, whose texture-oriented and high-frequency visual cues are partially smoothed by voxel-level aggregation.
Overall, these results indicate that CUS-GS improves semantic spatial coherence—particularly for high-level, language-oriented representations—without compromising overall multimodal alignment quality.

\begin{table*}[t]
    \centering
    \caption{\textbf{Quantitative Comparison on Downstream Task:} Our CUS-GS achieves comparable downstream performance to M3 with \emph{up to 5$\times$ fewer parameters}, demonstrating strong task generalization under compact model size. ``\# P.'' denotes number of parameters.}
    \resizebox{0.9\linewidth}{!}{
        \begin{tabular}{l|l|r|cccc|cccccc} \hline
            \multirow{2}{*}{\textbf{Dataset}} & \multirow{2}{*}{\textbf{Method}} & \multirow{2}{*}{\textbf{\# P.}} & \multicolumn{4}{c|}{\textbf{CLIP}} & \multicolumn{6}{c}{\textbf{SigLIP}} \\
            & & & mIoU & cIoU & AP50 & AP60 & I2T@1 & I2T@5 & I2T10 & T2I@1 & T2I@5 & T2I@10 \\
            \hline
            \multirow{4}{*}{Train}
            & Ground Truth &   - & 25.3 & 26.3 & 14.7 &  3.3 & 88.7 & 98.3 &  100 & 97.3 &  100 &  100 \\
            & F-3DGS~\cite{zhou_feature3dgs_2024_cvpr}       & 61M & 24.2 & 24.3 & 16.3 &  7.1 &  2.6 & 13.2 & 28.9 &  0.0 &  2.6 & 18.4 \\
            & M3~\cite{zou_m3spatial_2025_iclr}           & 35M & 25.4 & 26.5 & 19.6 & 12.5 & 55.2 & 84.2 & 92.1 & 52.6 & 84.2 & 92.1 \\
            & CUS-GS (Ours)         & 13M & 15.9 & 16.2 &  8.4 &  4.0 & 85.1 & 95.0 & 97.3 & 77.1 & 93.7 & 98.7 \\
            \hline
            \multirow{4}{*}{Playroom}
            & Ground Truth &   - & 25.6 & 24.2 &  9.6 & 3.0 & 92.9 & 98.7 &  100 & 92.0 & 98.2 & 98.1 \\
            & F-3DGS~\cite{zhou_feature3dgs_2024_cvpr}       & 61M & 23.8 & 21.4 & 11.9 & 3.0 & 79.3 & 96.6 &  6.6 & 31.0 & 79.3 & 89.7 \\
            & M3~\cite{zou_m3spatial_2025_iclr}           & 35M & 23.1 & 23.1 & 11.9 & 5.9 & 72.4 & 96.6 &  100 & 41.3 & 65.5 & 68.9 \\
            & CUS-3DGS (Ours)         & 6M & 15.5 & 16.2 &  4.8 & 2.1 & 37.3 & 74.2 & 81.8 & 58.2 & 84.8 & 90.2 \\
            \hline
        \end{tabular}
    }
    \label{table:downstream}
\end{table*}

\begin{figure*}
    \centering
    \includegraphics[width=\linewidth]{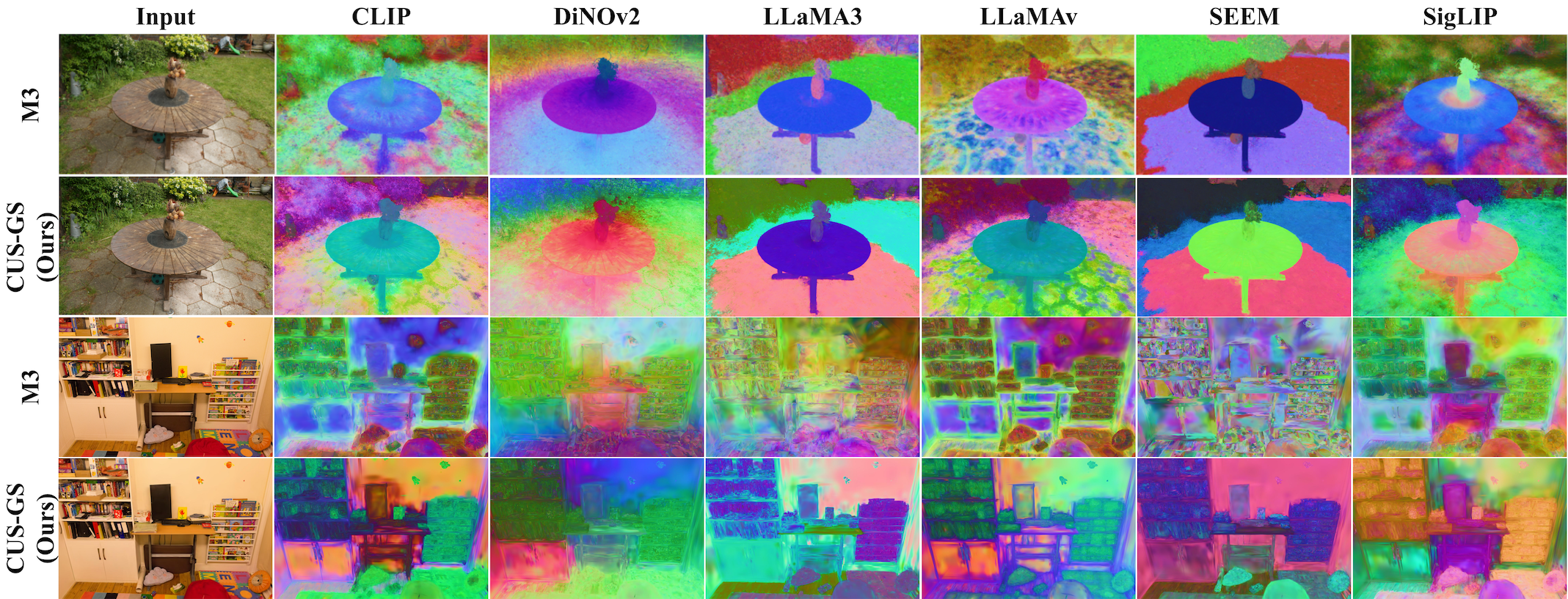}
    \caption{\textbf{Qualitative results:} Comparison of example feature maps reconstructed from different foundation models. Our  CUS-GS produces smoother and more spatially coherent representations than M3, with clearer structures in language-driven features (e.g., LLaMA3) and slightly smoother texture features (e.g., DINOv2).}
    \label{fig:qualitative}
\end{figure*}

\begin{table*}[t]
    \centering
    \caption{\textbf{Ablation Studies:} We analyze the effects of voxel size, query granularity, and feature-awareness (F.A.) ratio on model size, image rendering quality, and semantic feature alignment. }
    \resizebox{0.9\linewidth}{!}{
        \begin{tabular}{l|ccc|cc|cc|cc} \hline
            \multirow{2}{*}{\textbf{Dataset}} & \multicolumn{3}{c|}{\textbf{Config}} & \multicolumn{2}{c|}{\textbf{Model Size}} & \multicolumn{2}{c|}{\textbf{RBG Image}} & \multicolumn{2}{c}{\textbf{Semantic Features}} \\
            & Voxel Size & Query Granularity & F.A. Ratio & \#Param & Anchor & SSIM$\downarrow$ & PSNR$\downarrow$ & Avg. Cosine$\downarrow$ & Avg. $\ell_2$$\downarrow$ \\ 
            \hline
            \multicolumn{1}{c|}{\multirow{8}{*}{Train}} & 0.010 & per-anchor   & 2:8 &  13.4M & 128501 & 0.820 & 24.04 & 0.4006 & 9.4101 \\
            \cline{2-10}
            \multicolumn{1}{c|}{}                       & 0.050 & per-anchor   & 2:8 &   8.4M &  62639 & 0.752 & 23.05 & 0.4007 & 9.4114 \\
            \multicolumn{1}{c|}{}                       & 0.005 & per-anchor   & 2:8 &  14.3M & 141237 & 0.815 & 24.17 & 0.4006 & 9.4101 \\
            \cline{2-10}
            \multicolumn{1}{c|}{}                       & 0.010 & per-gaussian & 2:8 & 245.6M & 150624 & 0.845 & 25.27 & 0.3980 & 9.3958 \\
            \cline{2-10}
            \multicolumn{1}{c|}{}                       & 0.010 & per-anchor   & 8:2 &  13.4M & 129470 & 0.818 & 24.37 & 0.4006 & 9.4108 \\
            \multicolumn{1}{c|}{}                       & 0.010 & per-anchor   & 6:4 &  13.4M & 128955 & 0.830 & 24.79 & 0.4006 & 9.4101 \\
            \multicolumn{1}{c|}{}                       & 0.010 & per-anchor   & 4:6 &  13.4M & 129465 & 0.820 & 24.52 & 0.4006 & 9.4105 \\
            \hline
        \end{tabular}
    }
    \label{table:ablation}
\end{table*}

\autoref{table:downstream} reports the quantitative results on high-level downstream tasks. 
For CLIP-based semantic segmentation, CUS-GS shows a moderate drop compared with M3 and F-3DGS, 
which can be attributed both to the reduced model capacity (13–18M vs. 35–61M parameters) 
and the smoothing inherent to voxel-level aggregation.
Because CLIP features contain substantial global context and high-frequency variance, voxel aggregation suppresses local fluctuations and promotes spatial consistency, improving structural coherence while slightly weakening fine-grained semantic discrimination.
Nevertheless, CUS-GS maintains competitive segmentation accuracy under a significantly more compact configuration.  

For SigLIP-based image–text retrieval, the results exhibit a complementary trend: 
On the \emph{Train} scene, CUS-GS achieves performance close to the ground truth and clearly surpasses both baselines, 
indicating strong multimodal grounding in structured environments. 
In the more cluttered \emph{Playroom} scene, I2T performance decreases whereas T2I metrics remain higher than M3 and F-3DGS. 
Suggesting that CUS-GS preserves a more stable language-to-vision alignment, even under complex scene layouts. 

Overall, these results highlight that CUS-GS achieves a favorable balance between semantic consistency, structural quality, and model compactness, delivering competitive performance with a model size up to \textbf{5× smaller} than state-of-the-art counterparts.

\vspace{4pt}\noindent\textbf{Qualitative Results.} 
\autoref{fig:qualitative} shows qualitative comparisons of multimodal feature maps reconstructed from M3 and our CUS-GS. 
Overall, CUS-GS produces smoother and more spatially coherent representations.
For CLIP and SigLIP, the reconstructed features exhibit more consistent semantic regions, while LLaMA-based features show clearer spatial organization and improved alignment with scene structure. 
These observations align well with the quantitative results in \autoref{table:feat_align}, 
confirming that the voxel-structured design effectively preserves spatial hierarchy for language-driven semantics. 
For DINOv2, M3 retains slightly sharper texture cues, whereas CUS-GS appears more smoothed, which corresponds to the minor drop in cosine similarity reported in \autoref{table:feat_align}.
Overall, CUS-GS delivers more stable and spatially consistent multimodal representations across diverse foundation models.

\subsection{Ablation Studies}

Finally, we conduct three ablation studies in \autoref{table:ablation} to validate key design choices of CUS-GS. 
The first row reports the results of the \emph{Default Setup}, 
followed by ablations on \emph{Voxel Size} (rows 2–3), 
\emph{Query Granularity} (row 4), 
and \emph{Feature-Awareness Ratio} (rows 5–7).

\vspace{4pt}\noindent\textbf{Voxel Size.} As shown in \autoref{table:ablation}, the voxel size primarily influences the trade-off between model compactness and rendering quality. 
A smaller voxel size (0.005) leads to finer details but increases the number of anchors and parameters, 
while a larger voxel size (0.050) significantly reduces model size at the cost of degraded PSNR. 
We therefore choose $l=0.01$ as a balanced configuration that maintains high-quality rendering with moderate memory consumption.

\vspace{4pt}\noindent\textbf{Query Granularity.} Switching from per-anchor to per-Gaussian query substantially increases the model size (13.4M → 245.6M) 
with only marginal improvement in rendering quality (PSNR 24.04 → 25.27) and feature alignment. 
This indicates that hierarchical propagation from voxel anchors is sufficient to capture multimodal semantics without the need for redundant per-Gaussian queries. 

\vspace{4pt}\noindent\textbf{Feature-Awareness Ratio.} The feature-awareness ratio balances representation magnitude ($\lambda_{v,n}$) and learning dynamics ($\lambda_{v,g}$) in the pruning process. 
As shown in rows 5–7 of \autoref{table:ablation}, varying the ratio from 8:2 to 4:6 yields only marginal differences in SSIM and PSNR, 
indicating that the model is robust to moderate weighting changes. 
Larger $\lambda_{v,n}$ slightly improves PSNR but reduces compactness, while higher $\lambda_{v,g}$ maintains smaller anchor sets with comparable quality. 
We choose 2:8 as a balanced configuration that provides stable performance and efficient model size.

\section{Conclusion}
\label{sec:conclusion}

We have presented CUS-GS, the first Compact Unified Structured Gaussian Splatting framework for multimodal 3D scene representation. 
By bridging structural modeling and multimodal understanding, CUS-GS unifies spatial organization and semantic expressiveness within a single compact architecture. 
Through voxel-anchored latent features, hierarchical query adaptation, and feature-aware pruning, our method achieves competitive rendering quality and strong multimodal alignment while reducing model size by over 5× compared with existing approaches. 
Extensive experiments across diverse benchmarks demonstrate that CUS-GS effectively balances visual fidelity, semantic consistency, and compactness, 
showing clear advantages for both low-level reconstruction and high-level downstream tasks. 
We believe this work takes an important step toward unified, scalable, and semantically grounded 3D scene representations, 
and opens new directions for integrating foundation models into structured 3D learning.

{
    \small
    \bibliographystyle{ieeenat_fullname}
    \bibliography{main}
}

\clearpage
\setcounter{page}{1}
\maketitlesupplementary

\section{Overview}
\label{sec:overview}

In this supplementary material, we first provide additional experimental details in \cref{sec:exp_details}.
We then include extended qualitative results to further illustrate the capabilities of CUS-GS, 
including additional image rendering results in \cref{sec:rgb_results}, semantic feature rendering comparisons in \cref{sec:feature_results}, and
visualizations for downstream tasks in \cref{sec:downstream_results}.

\section{Additional Experimental Details}
\label{sec:exp_details}

In this section, we provide additional details on ground truth generation and evaluation procedure for the quantitative comparison in \autoref{table:downstream}. To ensure full consistency with the evaluation protocol of M3~\cite{zou_m3spatial_2025_iclr}, we adopt the same ground-truth construction pipeline for both CLIP~\cite{radford_clip_2021_icml} segmentation and SigLIP~\cite{zhai_siglip_2023_iccv} retrieval. Since SoM (semantic-SAM) and GPT-4o used in M3 are not publicly available, we replace them with fully open-source counterparts: SAM for mask generation and Qwen-VL for caption generation. This substitution does not change the mask granularity or the supervision structure, and all feature extraction and evaluation steps remain identical to M3.

For CLIP segmentation, we first apply SAM to the input RGB image to obtain region-level masks, which are then passed through Qwen-VL to obtain natural-language descriptions. 
During evaluation, the image feature maps are taken from the reconstructed multimodal representation, while the text embeddings are obtained by encoding the region descriptions with the CLIP text encoder.
Both features are normalized, and their dot-product similarity is computed to form similarity maps corresponding to each region description. These maps are upsampled to the resolution of the SAM masks and thresholded using Otsu’s method to produce predicted regions. We then compare the predictions with the SAM masks to compute mIoU, cIoU, and AP@0.5/0.6.

For SigLIP retrieval, we use Qwen-VL to generate a global caption for the target image, and then extract the image and text embeddings using the SigLIP vision and text encoders. We additionally encode COCO images and their captions with SigLIP to form a negative sample pool, which is combined with the target features to construct the retrieval set. During evaluation, we compute cosine similarities between the target image embedding and all text embeddings to obtain I2T@k, and likewise compute similarities between the target caption embedding and all image embeddings to obtain T2I@k.



\section{Image Rendering Results}
\label{sec:rgb_results}

\begin{figure*}
    \centering
    \includegraphics[width=0.95\linewidth]{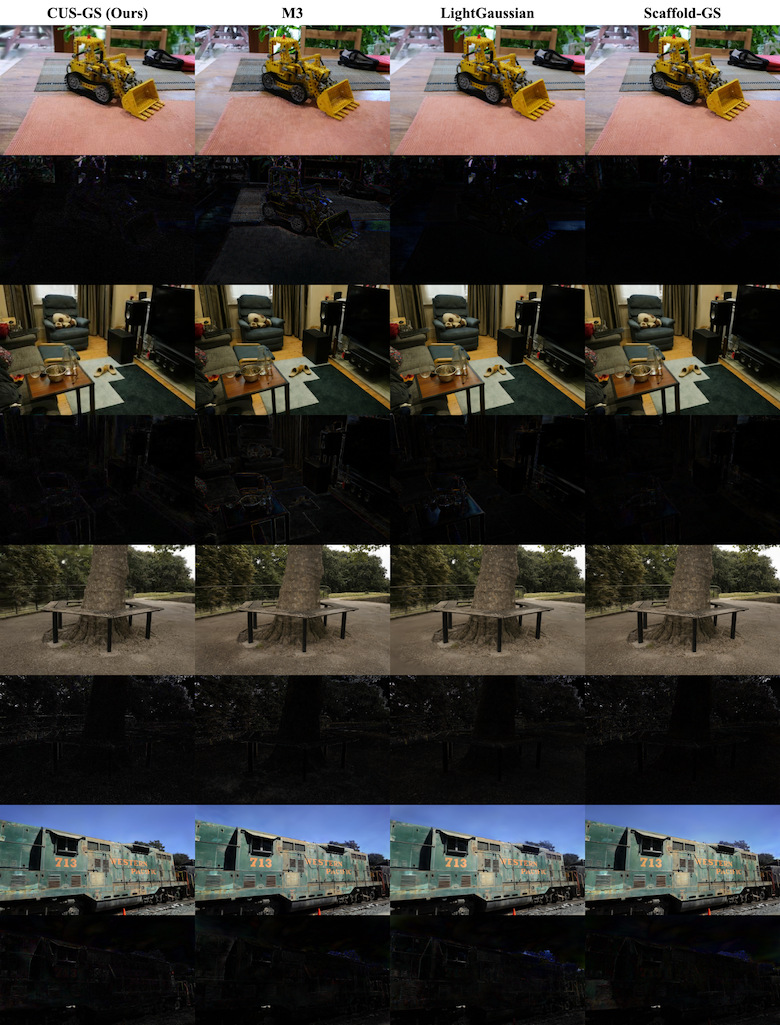}
    \caption{\textbf{Examples of the Image Rendering Results.} For each scene, the upper row shows the rendered image, and the lower row presents the residual between the rendered image and the ground truth image.}
    \label{fig:img_render}
\end{figure*}

\autoref{fig:img_render} presents qualitative comparisons of the image rendering results. While the rendered images across different methods appear visually similar at first glance, the residual maps reveal a much clearer distinction. Our CUS-GS consistently produces smaller and more uniformly distributed residuals, indicating more accurate color reconstruction and better alignment with ground-truth imagery. In contrast, both M3~\cite{zou_m3spatial_2025_iclr} (the only multimodal counterpart) and LightGaussian~\cite{fan_lightgaussian_2024_nips} exhibit noticeable structured errors, particularly around object boundaries and high-frequency regions, while Scaffold-GS~\cite{lu_scaffoldgs_2024_cvpr} shows moderate deviations.

\section{Semantic Feature Rendering Results}
\label{sec:feature_results}

\begin{figure*}[t]
    \centering
    \includegraphics[width=0.95\linewidth]{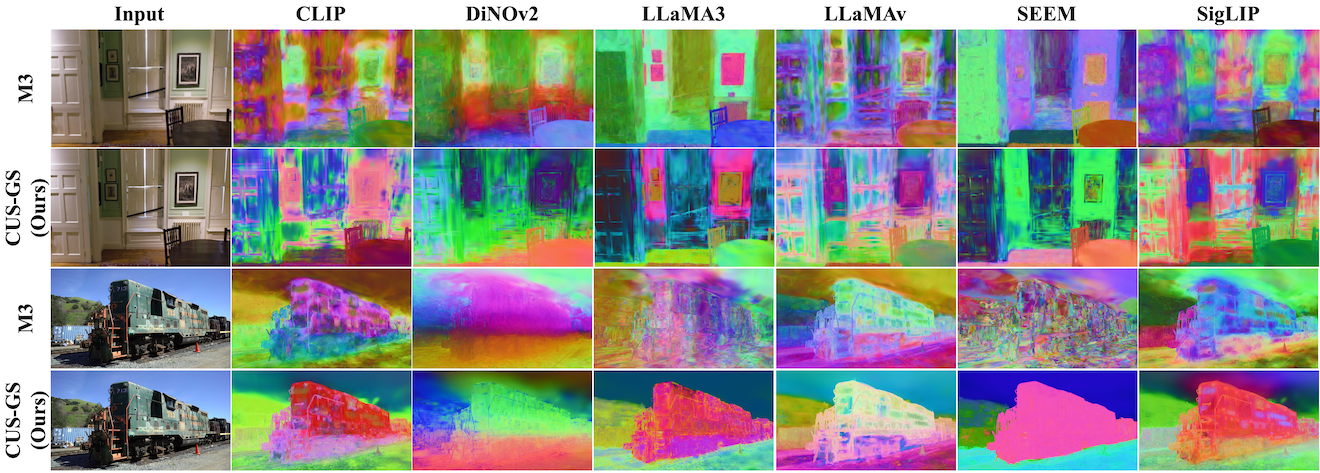}
    \caption{\textbf{Additional Feature Rendering Results.} These results provide complementary evidence that CUS-GS outperforms M3 in producing cleaner and more structured multimodal feature fields.}
    \label{fig:feature_render}
\end{figure*}

\autoref{fig:feature_render} presents additional semantic feature rendering results for the remaining two scenes, ``DrJohnson'' and ``Train,'' complementing the qualitative comparisons shown in the main text.
Across both scenes, CUS-GS consistently produces cleaner and more spatially coherent multimodal features, with CLIP and SigLIP exhibiting more stable semantic regions and LLaMA-based features showing clearer structural organization. As observed previously, DINOv2 features from CUS-GS appear slightly smoother than those from M3, corresponding to the minor alignment trade-off discussed in \autoref{table:feat_align}.
Overall, these additional visualizations further confirm the robustness and spatial consistency of the multimodal representations rendered by CUS-GS.

\section{Downstream Tasks Visualization}
\label{sec:downstream_results}

\begin{figure*}
    \centering
    \includegraphics[width=0.95\linewidth]{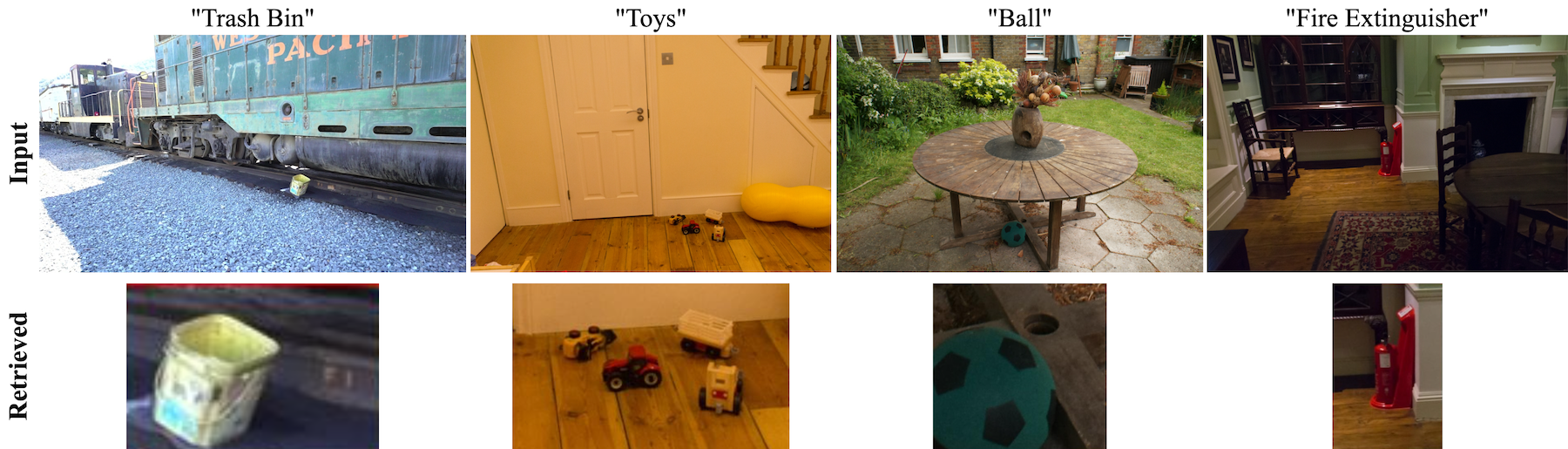}
    \caption{\textbf{Examples of Retrieval Tasks.} Given text queries (“Trash Bin”, “Toys”, “Ball”, “Fire Extinguisher”), we show the top-1 retrievals using CUS-GS’s reconstructed features. The top row displays example views from the corresponding scenes, and the bottom row demonstrates the retrieved target, successfully locating the correct object instances.}
    \label{fig:retrieval_example}
\end{figure*}

\begin{figure*}
    \centering
    \includegraphics[width=0.95\linewidth]{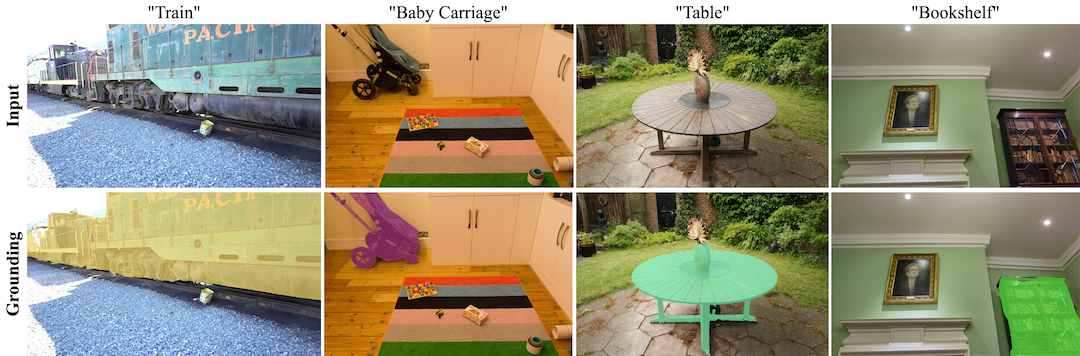}
    \caption{\textbf{Examples of Grounding Tasks.} The upper row shows the input view and text prompt, and the lower row presents the object mask.}
    \label{fig:grounding_example}
\end{figure*}

\begin{figure*}
    \centering
    \includegraphics[width=0.95\linewidth]{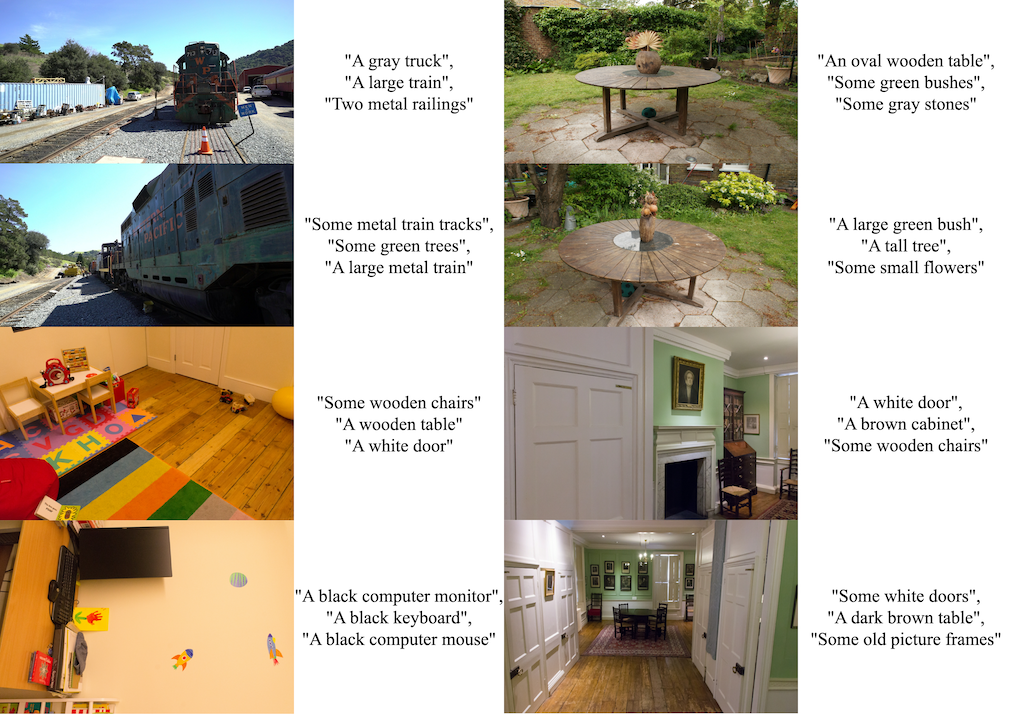}
    \caption{\textbf{Examples of Caption Tasks.} The lefter column shows the input view and the righter column demonstrates the generated caption.}
    \label{fig:caption_example}
\end{figure*}

In this section, we provide additional visualizations for the downstream tasks—retrieval, grounding, and caption—as shown in \autoref{fig:retrieval_example}, \autoref{fig:grounding_example}, and \autoref{fig:caption_example}. These examples illustrate how the multimodal features reconstructed by CUS-GS can be directly used in practical semantic reasoning tasks through the decoders of their corresponding foundation models. The results show that the rendered features remain well aligned with the native feature spaces of these models, enabling them to produce meaningful and coherent outputs without any task-specific adaptation. This further verifies that CUS-GS preserves not only semantic consistency but also functional usability across diverse downstream scenarios.

\end{document}